\documentclass[final]{cvpr}
\usepackage{subfig}
\usepackage{times}
\usepackage{epsfig}
\usepackage{graphicx}
\usepackage{amsmath}
\usepackage{amssymb}
\usepackage{multirow}
\usepackage{multicol}
\newcommand{\tablestyle}[2]{\setlength{\tabcolsep}{#1}\renewcommand{\arraystretch}{#2}\centering\small}
\newlength\savewidth\newcommand\shline{\noalign{\global\savewidth\arrayrulewidth
  \global\arrayrulewidth 1pt}\hline\noalign{\global\arrayrulewidth\savewidth}}

\usepackage[pagebackref=true,breaklinks=true,colorlinks,bookmarks=false]{hyperref}

\def\cvprPaperID{10} 
\def\confYear{CVPR 2021}

\begin{document}

\title{A Stronger Baseline for Ego-Centric Action Detection}

\author{Zhiwu Qing$^{1,4\dag}$ \quad Ziyuan Huang$^{2,4\dag}$\quad  Xiang Wang$^{1,4}$ \quad Yutong Feng$^{3,4}$ \quad Shiwei Zhang$^{4*}$ \\ Jianwen Jiang$^4$  \quad Mingqian Tang$^4$
\quad  Changxin Gao$^1$ \quad Marcelo H. Ang Jr$^2$   \quad Nong Sang$^{1*}$
\\
$^1$Key Laboratory of Image Processing and Intelligent Control \\ School of Artificial Intelligence and Automation, Huazhong University of Science and Technology\\
$^2$ARC, National University of Singapore\\
$^3$Tsinghua University \quad
$^4$Alibaba Group\\
{\tt\small \{qzw, wxiang, cgao, nsang\}@hust.edu.cn}\\
{\tt\small ziyuan.huang@u.nus.edu, mpeangh@nus.edu.sg}\\
{\tt\small fyt19@mails.tsinghua.edu.cn}\\
{\tt\small \{zhangjin.zsw, jianwen.jjw, mingqian.tmq\}@alibaba-inc.com}
}

\maketitle

\let\thefootnote\relax\footnotetext{$\dag$ Equal Contribution.}
\let\thefootnote\relax\footnotetext{$*$ Corresponding authors.}
\let\thefootnote\relax\footnotetext{This work is done when Z. Qing, Z. Huang, X. Wang and Y. Feng are interns at Alibaba Group.}

\begin{abstract}
   This technical report analyzes an egocentric video action detection method we used in the 2021 EPIC-KITCHENS-100 competition hosted in CVPR2021 Workshop.
   The goal of our task is to locate the start time and the end time of the action in the long untrimmed video, and predict action category.
   We adopt sliding window strategy to generate proposals, which can better adapt to short-duration actions. In addition, we show that classification and proposals are conflict in the same network. The separation of the two tasks boost the detection performance with high efficiency.
   By simply employing these strategy, we achieved 16.10\% performance on the test set of EPIC-KITCHENS-100 Action Detection challenge using a single model, surpassing the baseline method by 11.7\% in terms of average mAP.
   
\end{abstract}

\section{Introduction}
 Temporal action detection is a challenging task, especially for the EPIC-KITCHENS-100 dataset~\cite{damen2020ek100}, where (a) most actions spans a short period, compared to the duration of the original untrimmed videos and (b) consistently altering action categories under the same background environment requires the network to have the ability to look for fine-grained features and discriminate complicated spatio-temporal interactions.
 To alleviate these issues, we propose the following strategies: 
 (a) We use sliding windows to restrict the length of the input untrimmed videos for each video clip that is to be evaluated. This ensures that enough features are assigned to the short action segment candidates, which will be otherwise overwhelmed by the features from other segments in a long video that is simply normalized.
 The possibility is also increased that the length of the potential action segments can be matched to the pre-defined temporal anchors.
 (b) For more accurate verb and noun classifications, pre-trained backbones are employed for the classification of each video clip in the long videos. 
Additionally, we noticed an optimization conflict for proposal evaluation and classification, where the performances of both tasks drop drastically when a joint head is used to perform both tasks. Hence, we propose to use separate heads for evaluating the proposals as well as performing the classifications.

\begin{figure*}
\begin{center}
\includegraphics[width=18cm]{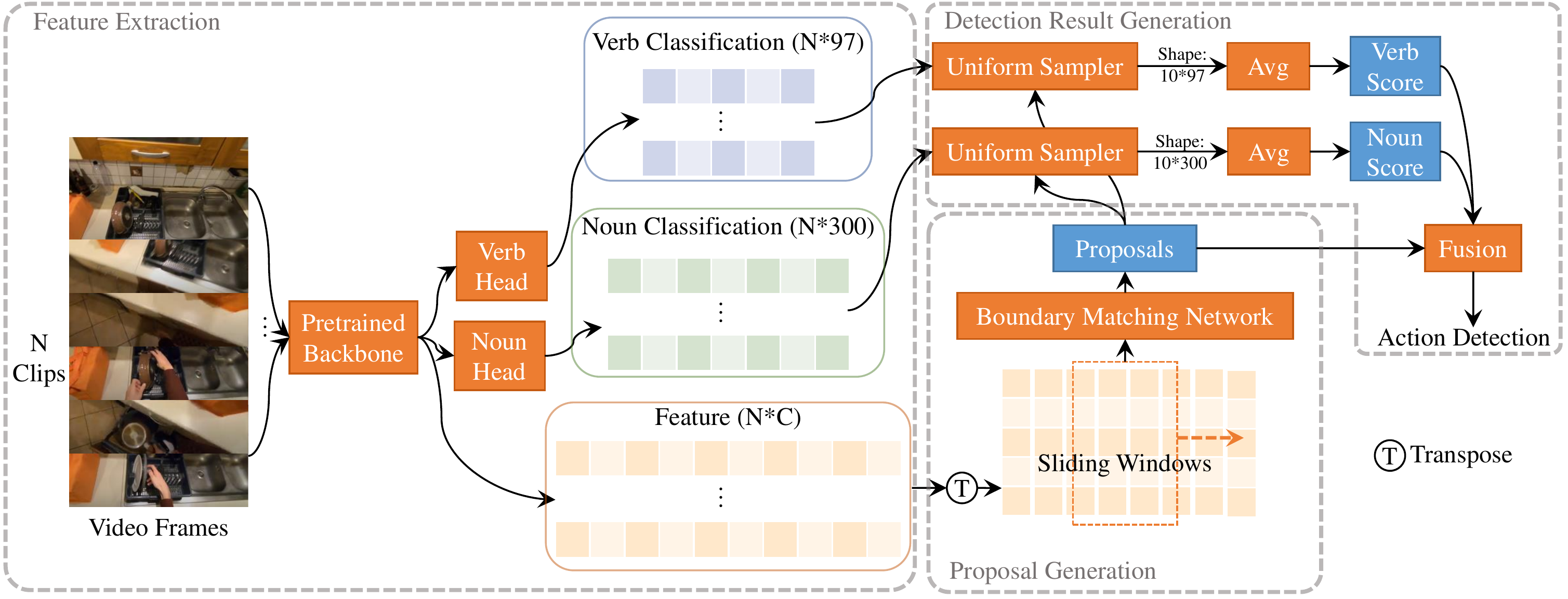}
\end{center}
   \caption{\textbf{The overall framework of our approach.} In feature extraction process, the input videos are divided into N clips, which are fed to the pre-trained backbone to extract features as well as verb and noun predictions. In the proposal generation stage, the sliding windows are uniformly distributed along temporal dimension, and clip-level features covered by each sliding window are fed to the Boundary Matching Network to generate proposals. In detection generation stage, the classification results for each proposal are sampled from classification scores yielded by the pre-trained backbone. Finally, the verb and noun predictions are fused with proposals to generate detection result with action predictions.}
\label{fig:short}
\end{figure*}

\section{Our Approach}
The overall architecture of our approach is visualized in Figure~\ref{fig:short}. The general process can be divided into four steps, respectively, \textbf{(i)} the pre-training of the classification models, \textbf{(ii)} feature extraction process, \textbf{(iii)} proposal generation process as well as the \textbf{(iv)} detection result generation process. We will discuss all the four stages one by one in the following sections. 
\subsection{Pre-train of Classification Models}
Transfer learning is an important measure to improve the generalization ability of the model.
Supervised training~\cite{qiu2017learning_p3d,tran2018closer_look,carreira2017quo_inception_3d,xie1712rethinking,tran2019csn,feichtenhofer2019slowfast} as well as unsupervised ones~\cite{huang2021mosi,han2020coclr,miech2020milnce} are two mainstream pre-training strategy. Although the latter strategy can leverage a larger set of data, leading to a more generalized representation, supervised pre-training utilizes training data more efficiently and effectively. Therefore, we adopt the former strategy.
Recently, Transformer-Based methods have shown great potential in image recognition~\cite{dosovitskiy2020vit, zhou2021deepvit} and video understanding~\cite{arnab2021vivit,bertasius2021timesformer}. We employ ViViT~\cite{arnab2021vivit} and CSN~\cite{tran2019csn} as our backbone for comparison and first pre-train it on Kinetics700~\cite{carreira2019k700} dataset, mostly following the training recipes in DeepViT~\cite{zhou2021deepvit}.  And then the pre-trained backbone is fine-tuned on EPIC-KITCHENS-100 dataset with verb head and noun head. In the fine-tuning stage, the FPS of the videos are normalized to 60, we sample 32 frames with sampling rate of 2 for each clip. The training details of the backbone models can be referred to our Action Recognition report~\cite{huang2021epic}.

\begin{table*}[t]
    \centering
\subfloat[\textbf{Action detection results for Action.}\label{tab:action_map}]{
\tablestyle{4pt}{1.0}
\small
\begin{tabular}{c|c|cccccc|cccccc}
    \multirow{2}{*}{Feature Backbone} &\multirow{2}{*}{Classification} & \multicolumn{6}{c|}{mAP(Val) for Action}& \multicolumn{6}{c}{mAP(Test) for Action} \\
     &  & @0.1 & @0.2 & @0.3 & @0.4 & @0.5& Avg & @0.1 & @0.2 & @0.3 & @0.4 & @0.5& Avg \\
    \shline
    ViViT~\cite{arnab2021vivit} & BMN~\cite{lin2019bmn} & 6.84 & 6.01 & 5.28 & 4.42&3.26 & 5.16 &5.86 &5.33 &4.67 &4.03 &3.03 & 4.59 \\
    CSN~\cite{tran2019csn}& BMN~\cite{lin2019bmn} &7.30 &7.01 & 6.56&6.05 & 5.08&  6.40 &- & -&- &- & -&  - \\
    ViViT~\cite{arnab2021vivit} & CSN~\cite{tran2019csn} &13.90 & 13.23 &11.98 &10.48 &8.80 &  11.68 & 13.08 &11.97 &10.84 &9.56 &8.00 &  10.69 \\
    ViViT~\cite{arnab2021vivit} & ViViT~\cite{arnab2021vivit} &\textbf{21.14} &\textbf{20.10} &\textbf{19.02} &\textbf{17.32} &\textbf{15.11} &  \textbf{18.53} & \textbf{18.76} & \textbf{17.73} & \textbf{16.26} &\textbf{14.91} &\textbf{12.87} &  \textbf{16.11} \\

\end{tabular}
}

\subfloat[\textbf{Action detection results for Verb.}\label{tab:verb_map}]{
\tablestyle{4pt}{1.0}
\small
\begin{tabular}{c|c|cccccc|cccccc}
    \multirow{2}{*}{Feature Backbone} &\multirow{2}{*}{Classification} & \multicolumn{6}{c|}{mAP(Val) for Verb}& \multicolumn{6}{c}{mAP(Test) for Verb} \\
     &  & @0.1 & @0.2 & @0.3 & @0.4 & @0.5& Avg & @0.1 & @0.2 & @0.3 & @0.4 & @0.5& Avg \\
    \shline
    ViViT~\cite{arnab2021vivit} & BMN~\cite{lin2019bmn} & 11.32&10.07 & 8.64& 6.73& 5.00 & 8.35 &10.07 &9.41 &8.29 & 6.63 & 4.80 & 7.84 \\
    CSN~\cite{tran2019csn}& BMN~\cite{lin2019bmn} & 12.89& 12.39&11.55 &10.42 &8.09 & 11.06  & -& -& -&- &- &  - \\
    ViViT~\cite{arnab2021vivit} & CSN~\cite{tran2019csn} & 16.57 & 15.56 & 14.10 &12.12 &10.21 & 13.71  &17.58 &15.91 &14.21 &12.23 & 9.73& 13.93  \\
    ViViT~\cite{arnab2021vivit} & ViViT~\cite{arnab2021vivit} & \textbf{22.92}&\textbf{21.86} &\textbf{20.89} &\textbf{18.33} & \textbf{15.66} &  \textbf{19.93} & \textbf{22.77} & \textbf{22.01} & \textbf{19.63} & \textbf{17.81} &\textbf{14.65} &  \textbf{19.37} \\
\end{tabular}
}

\subfloat[\textbf{Action detection results for Noun.}\label{tab:noun_map}]{
\tablestyle{4pt}{1.0}
\small
\begin{tabular}{c|c|cccccc|cccccc}
    \multirow{2}{*}{Feature Backbone} &\multirow{2}{*}{Classification}  & \multicolumn{6}{c|}{mAP(Val) for Noun}& \multicolumn{6}{c}{mAP(Test) for Noun} \\
    ~ & ~ & @0.1 & @0.2 & @0.3 & @0.4 & @0.5& Avg & @0.1 & @0.2 & @0.3 & @0.4 & @0.5& Avg \\
    \shline
    ViViT~\cite{arnab2021vivit} & BMN~\cite{lin2019bmn} &9.70 &8.35 & 7.21 & 5.77 & 4.08 & 7.02 &9.76 &8.67 &7.43 &6.02 &4.19 & 7.22 \\
    CSN~\cite{tran2019csn}& BMN~\cite{lin2019bmn} & 11.00& 10.34& 9.46&8.29 &6.71&  9.16 &- & -& -& -&- &  - \\
    ViViT~\cite{arnab2021vivit} & CSN~\cite{tran2019csn} & 18.47&17.21 &15.56 &13.38 &10.58 & 15.04  & 19.46& 17.79&15.87 & 13.62&10.90 &  15.53 \\
    ViViT~\cite{arnab2021vivit} & ViViT~\cite{arnab2021vivit} &\textbf{30.09} &\textbf{27.59} &\textbf{25.81} & \textbf{22.80} & \textbf{19.26} &  \textbf{25.11} & \textbf{26.44} & \textbf{24.55} & \textbf{22.30} & \textbf{19.82} &\textbf{16.25} &  \textbf{21.87} \\

\end{tabular}
}
    \caption{\textbf{Action detection results on EPIC-KITCHENS-100 dataset. }Features are extracted by the backbone in the feature backbone column. BMN in the classification column indicates that two classification heads are added upon the BMN feature to perform verb and noun predictions, while CSN and ViViT in the classification column indicates that we directly sample the prediction results temporally to obtain the final classification. The predictions results of CSN and ViViT are saved during the feature extraction process.}
    \label{tab:det_results}
\end{table*}

\subsection{Feature Extractor}
Limited by the GPU memory, the raw frames cannot be directly fed to the backbone. 
Because of the limited GPU memory, it is impossible to put the whole video to the computing device. 
Therefore, multiple high-dimensional feature vectors are extracted from the untrimmed video using the pre-trained backbones, which will be further used in the later stage to generate action proposals.
For the feature extraction process, we mostly follow the common setting in the temporal action detection community~\cite{lin2019bmn, lin2018bsn,lin2020dbg, xu2020gtad,gao2017turn,buch2017sst,bai2020bcgnn,zhang2019glnet,qing2021tca,qing2020tfn,wang2020cbr,anetava2018,wang2021self}.
Specifically, given the number of frames $l$ in a video, we split the video according to a fixed stride $\delta$ between consecutive video clips. Hence, the whole video is split into $N$ clips, where $N=l/\delta$. In our experiments, the value of $\delta$ is set to 16. 
It is worth noting that, besides the feature vectors, the verb and noun predictions are saved at the same time for each clip when extracting its features.

\subsection{Generation of Proposals}
In our observations, different from mainstream action detection datasets~\cite{caba2015activitynet,zhao2019hacs}, 98.15\% of the duration of the ground truth action segments are less than 20s. However, the average duration of the videos is up to 512.43s, which results in an extremely low percentage that the action segment candidate accounts for in the entire video.
Therefore, we propose to generate action proposals within sliding windows. 
For each sliding window, we include features for 200 video clips. Because the interval between two consecutive video clips are 16 frames, which can be converted to 0.2667s in a 60-fps video, each sliding window contains contextual information lasting around 53.33s. 
The time interval between each sliding window is half of the size of one sliding window, which is 26.67 seconds. 
To ensure that at least one sliding window will cover the whole action segment candidate, we limit the maximum length of the potential action segment to be 26.67 seconds.

With sliding windows, Boundary Matching Network(BMN)~\cite{lin2019bmn} is employed to generate accurate proposals. 
Given the clip-level features $\mathbf{x}\in\mathbb{R}^{N\times C}$, BMN extracts candicate-level features for each potential action segments by matrix multiplication. 
The resultant features are then scored according to their IoU with the ground truth.
We discard the scores from the Temporal Evaluation Module(TEM) and only retain the scores from the regression map and the classification map in Proposal Evaluation Module(PEM), as the inclusion of TEM actually hurt the performance. For more details obout PEM and TEM, please refer to BMN~\cite{lin2019bmn}.
Since there is redundancy in the proposals generated by BMN, Soft-NMS~\cite{bodla2017softnms} is applied to remove the redundant proposals. The hyperparameters in Soft-NMS are set to 0.25, 0.9 and 0.4 for low threshold, high threshold and alpha, respectively.
In training, we utilize AdamW~\cite{loshchilov2017adamw} as optimizer and set learning rate to 0.002. The model is trained for 10 epochs with cosine learning rate schedule.

\subsection{Generation of Detection Results}

To detect actions, it would be convenient if the BMN~\cite{lin2019bmn} can directly output both scores for the candidate proposals as well as the predictions of the verb and noun category for the corresponding proposals.
However, we observe that when we use the candidate-level features extracted by the BMN network to perform both proposal evaluation task and classification, the performance is terrible. 
We suspect that there is some optimization conflict in the classification and the proposal evaluation tasks. 

Since the accuracy of the feature extraction backbone ViViT (pre-trained on Kinetics700 and fine-tuned for EPIC-KITCHENS-100 action recognition task) can achieve 47.4\% with 3$\times$10 views in the validation set, we directly use its classification predictions.
Specifically, we save all the predictions during the feature extraction process as we have mentioned before. 
Empirically, we show in Table~\ref{tab:det_results} that, when we use ViViT~\cite{arnab2021vivit} or CSN~\cite{tran2019csn} features as the clip-level features and the candidate-level features of BMN to perform classification, the performance is worse than simply using the classification results generated directly by ViViT or CSN.
Furthermore, when ViViT classification is used, a 6.85\% performance improvement is observed over the CSN classification results. This is mainly due to the higher accuracy of ViViT in action recognition task. In our experiments, we do not use any ensemble strategy, which provides a simple and strong baseline for EPIC-KITCHENS-100 dataset.

To generate detection results, for each proposal generated by BMN, we sample the classification results in time range covered by the proposal with 10 uniform temporal location. The sampled classification results are averaged to get the prediction of respectively verb and noun for the proposal. 
Finally, the action detection results are obtained by fusing verb, noun scores and proposals.


\section{Conclusion}
In this report, we propose a stronger baseline for Ego-Centric Action Detection. We adopt a sliding window strategy to alleviate the problem that the temporal duration of proposals is too short to be detected difficultly. In addition, we also found that the conflict when the classification task and the proposal task coexist in the same network. Separating the two has significantly improved the performance. These two problems are inevitable in EPIC-KITCHENS-100 temporal action detection, how to solve these problems elegantly is still worthy of further study.

\section{Acknowledgment}
This work is supported by the National Natural Science Foundation of China under grant 61871435 and the Fundamental Research Funds for the Central Universities no. 2019kfyXKJC024, the Agency for Science, Technology and Research (A*STAR) under its AME Programmatic Funding Scheme (Project A18A2b0046) and by Alibaba Group through Alibaba Research Intern Program.

{\small
\bibliographystyle{ieee_fullname}
\bibliography{egbib}

\begin{thebibliography}{10}\itemsep=-1pt

\bibitem{arnab2021vivit}
Anurag Arnab, Mostafa Dehghani, Georg Heigold, Chen Sun, Mario Lu{\v{c}}i{\'c},
  and Cordelia Schmid.
\newblock Vivit: A video vision transformer.
\newblock {\em arXiv preprint arXiv:2103.15691}, 2021.

\bibitem{bai2020bcgnn}
Yueran Bai, Yingying Wang, Yunhai Tong, Yang Yang, Qiyue Liu, and Junhui Liu.
\newblock Boundary content graph neural network for temporal action proposal
  generation.
\newblock {\em arXiv preprint arXiv:2008.01432}, 2020.

\bibitem{bertasius2021timesformer}
Gedas Bertasius, Heng Wang, and Lorenzo Torresani.
\newblock Is space-time attention all you need for video understanding?
\newblock {\em arXiv preprint arXiv:2102.05095}, 2021.

\bibitem{bodla2017softnms}
Navaneeth Bodla, Bharat Singh, Rama Chellappa, and Larry~S Davis.
\newblock Soft-nms--improving object detection with one line of code.
\newblock In {\em Proceedings of the IEEE international conference on computer
  vision}, pages 5561--5569, 2017.

\bibitem{buch2017sst}
Shyamal Buch, Victor Escorcia, Chuanqi Shen, Bernard Ghanem, and Juan
  Carlos~Niebles.
\newblock Sst: Single-stream temporal action proposals.
\newblock In {\em Proceedings of the IEEE conference on Computer Vision and
  Pattern Recognition}, pages 2911--2920, 2017.

\bibitem{caba2015activitynet}
Fabian Caba~Heilbron, Victor Escorcia, Bernard Ghanem, and Juan Carlos~Niebles.
\newblock Activitynet: A large-scale video benchmark for human activity
  understanding.
\newblock In {\em Proceedings of the ieee conference on computer vision and
  pattern recognition}, pages 961--970, 2015.

\bibitem{carreira2019k700}
Joao Carreira, Eric Noland, Chloe Hillier, and Andrew Zisserman.
\newblock A short note on the kinetics-700 human action dataset.
\newblock {\em arXiv preprint arXiv:1907.06987}, 2019.

\bibitem{carreira2017quo_inception_3d}
Joao Carreira and Andrew Zisserman.
\newblock Quo vadis, action recognition? a new model and the kinetics dataset.
\newblock In {\em IEEE Conf. Comput. Vis. Pattern Recog.}, pages 6299--6308,
  2017.

\bibitem{damen2020ek100}
Dima Damen, Hazel Doughty, Giovanni~Maria Farinella, Antonino Furnari,
  Evangelos Kazakos, Jian Ma, Davide Moltisanti, Jonathan Munro, Toby Perrett,
  Will Price, et~al.
\newblock Rescaling egocentric vision.
\newblock {\em arXiv preprint arXiv:2006.13256}, 2020.

\bibitem{dosovitskiy2020vit}
Alexey Dosovitskiy, Lucas Beyer, Alexander Kolesnikov, Dirk Weissenborn,
  Xiaohua Zhai, Thomas Unterthiner, Mostafa Dehghani, Matthias Minderer, Georg
  Heigold, Sylvain Gelly, et~al.
\newblock An image is worth 16x16 words: Transformers for image recognition at
  scale.
\newblock {\em arXiv preprint arXiv:2010.11929}, 2020.

\bibitem{feichtenhofer2019slowfast}
Christoph Feichtenhofer, Haoqi Fan, Jitendra Malik, and Kaiming He.
\newblock Slowfast networks for video recognition.
\newblock In {\em Proceedings of the IEEE/CVF International Conference on
  Computer Vision}, pages 6202--6211, 2019.

\bibitem{gao2017turn}
Jiyang Gao, Zhenheng Yang, Kan Chen, Chen Sun, and Ram Nevatia.
\newblock Turn tap: Temporal unit regression network for temporal action
  proposals.
\newblock In {\em Proceedings of the IEEE international conference on computer
  vision}, pages 3628--3636, 2017.

\bibitem{han2020coclr}
Tengda Han, Weidi Xie, and Andrew Zisserman.
\newblock Self-supervised co-training for video representation learning.
\newblock {\em arXiv preprint arXiv:2010.09709}, 2020.

\bibitem{huang2021epic}
Ziyuan Huang, Zhiwu Qing, Xiang Wang, Yutong Feng, Shiwei Zhang, Jianwen Jiang,
  Zhurong Xia, Mingqian Tang, Nong Sang, and Marcelo Ang.
\newblock Towards training stronger video vision transformers for
  epic-kitchens-100 action recognition.
\newblock {\em arXiv preprint arXiv:2106.05058}, 2021.

\bibitem{huang2021mosi}
Ziyuan Huang, Shiwei Zhang, Jianwen Jiang, Mingqian Tang, Rong Jin, and Marcelo
  Ang.
\newblock Self-supervised motion learning from static images.
\newblock {\em arXiv preprint arXiv:2104.00240}, 2021.

\bibitem{anetava2018}
Jianwen Jiang, Yu Cao, Lin Song, SZY Li, Z Xu, C~Gan Q~Wu, C Zhang, and G Yu.
\newblock Human centric spatio-temporal action localization.
\newblock {\em ActivityNet Workshop on CVPR}, 2018.

\bibitem{lin2020dbg}
Chuming Lin, Jian Li, Yabiao Wang, Ying Tai, Donghao Luo, Zhipeng Cui, Chengjie
  Wang, Jilin Li, Feiyue Huang, and Rongrong Ji.
\newblock Fast learning of temporal action proposal via dense boundary
  generator.
\newblock In {\em AAAI}, pages 11499--11506, 2020.

\bibitem{lin2019bmn}
Tianwei Lin, Xiao Liu, Xin Li, Errui Ding, and Shilei Wen.
\newblock Bmn: Boundary-matching network for temporal action proposal
  generation.
\newblock In {\em Proceedings of the IEEE International Conference on Computer
  Vision}, pages 3889--3898, 2019.

\bibitem{lin2018bsn}
Tianwei Lin, Xu Zhao, Haisheng Su, Chongjing Wang, and Ming Yang.
\newblock Bsn: Boundary sensitive network for temporal action proposal
  generation.
\newblock In {\em Proceedings of the European Conference on Computer Vision
  (ECCV)}, pages 3--19, 2018.

\bibitem{loshchilov2017adamw}
Ilya Loshchilov and Frank Hutter.
\newblock Decoupled weight decay regularization.
\newblock {\em arXiv preprint arXiv:1711.05101}, 2017.

\bibitem{miech2020milnce}
Antoine Miech, Jean-Baptiste Alayrac, Lucas Smaira, Ivan Laptev, Josef Sivic,
  and Andrew Zisserman.
\newblock End-to-end learning of visual representations from uncurated
  instructional videos.
\newblock In {\em Proceedings of the IEEE/CVF Conference on Computer Vision and
  Pattern Recognition}, pages 9879--9889, 2020.

\bibitem{qing2021tca}
Zhiwu Qing, Haisheng Su, Weihao Gan, Dongliang Wang, Wei Wu, Xiang Wang, Yu
  Qiao, Junjie Yan, Changxin Gao, and Nong Sang.
\newblock Temporal context aggregation network for temporal action proposal
  refinement.
\newblock {\em arXiv preprint arXiv:2103.13141}, 2021.

\bibitem{qing2020tfn}
Zhiwu Qing, Xiang Wang, Yongpeng Sang, Changxin Gao, Shiwei Zhang, and Nong
  Sang.
\newblock Temporal fusion network for temporal action localization: Submission
  to activitynet challenge 2020 (task e).
\newblock {\em arXiv preprint arXiv:2006.07520}, 2020.

\bibitem{qiu2017learning_p3d}
Zhaofan Qiu, Ting Yao, and Tao Mei.
\newblock Learning spatio-temporal representation with pseudo-3d residual
  networks.
\newblock In {\em Int. Conf. Comput. Vis.}, pages 5533--5541, 2017.

\bibitem{tran2019csn}
Du Tran, Heng Wang, Lorenzo Torresani, and Matt Feiszli.
\newblock Video classification with channel-separated convolutional networks.
\newblock In {\em Proceedings of the IEEE/CVF International Conference on
  Computer Vision}, pages 5552--5561, 2019.

\bibitem{tran2018closer_look}
Du Tran, Heng Wang, Lorenzo Torresani, Jamie Ray, Yann LeCun, and Manohar
  Paluri.
\newblock A closer look at spatiotemporal convolutions for action recognition.
\newblock In {\em IEEE Conf. Comput. Vis. Pattern Recog.}, pages 6450--6459,
  2018.

\bibitem{wang2020cbr}
Xiang Wang, Baiteng Ma, Zhiwu Qing, Yongpeng Sang, Changxin Gao, Shiwei Zhang,
  and Nong Sang.
\newblock Cbr-net: Cascade boundary refinement network for action detection:
  Submission to activitynet challenge 2020 (task 1).
\newblock {\em arXiv preprint arXiv:2006.07526}, 2020.

\bibitem{wang2021self}
Xiang Wang, Shiwei Zhang, Zhiwu Qing, Yuanjie Shao, Changxin Gao, and Nong
  Sang.
\newblock Self-supervised learning for semi-supervised temporal action
  proposal.
\newblock {\em arXiv preprint arXiv:2104.03214}, 2021.

\bibitem{xie1712rethinking}
S Xie, C Sun, J Huang, Z Tu, and K Murphy.
\newblock Rethinking spatiotemporal feature learning for video understanding
  (2017). arxiv preprint.
\newblock {\em arXiv preprint arXiv:1712.04851}, 2017.

\bibitem{xu2020gtad}
Mengmeng Xu, Chen Zhao, David~S Rojas, Ali Thabet, and Bernard Ghanem.
\newblock G-tad: Sub-graph localization for temporal action detection.
\newblock In {\em Proceedings of the IEEE/CVF Conference on Computer Vision and
  Pattern Recognition}, pages 10156--10165, 2020.

\bibitem{zhang2019glnet}
Shiwei Zhang, Lin Song, Changxin Gao, and Nong Sang.
\newblock Glnet: Global local network for weakly supervised action
  localization.
\newblock {\em IEEE Transactions on Multimedia}, 22(10):2610--2622, 2019.

\bibitem{zhao2019hacs}
Hang Zhao, Antonio Torralba, Lorenzo Torresani, and Zhicheng Yan.
\newblock Hacs: Human action clips and segments dataset for recognition and
  temporal localization.
\newblock In {\em Proceedings of the IEEE/CVF International Conference on
  Computer Vision}, pages 8668--8678, 2019.

\bibitem{zhou2021deepvit}
Daquan Zhou, Bingyi Kang, Xiaojie Jin, Linjie Yang, Xiaochen Lian, Zihang
  Jiang, Qibin Hou, and Jiashi Feng.
\newblock Deepvit: Towards deeper vision transformer.
\newblock {\em arXiv preprint arXiv:2103.11886}, 2021.

\end{thebibliography}
}

\end{document}